\documentclass[runningheads]{llncs}
\usepackage[T1]{fontenc}
\usepackage{graphicx}
\usepackage{pifont} 
\usepackage{CJKutf8}
\usepackage{paralist}
\newcommand{\resultone}[1]{\colorbox{green!15}{#1}}
\newcommand{\resulttwo}[1]{\colorbox{cyan!15}{#1}}
\newcommand{\resultthird}[1]{\colorbox{yellow!15}{#1}}
\newcommand{\bench}{\textsc{H2HTalk}}
\newcommand{\numModels}{\textsc{50}}

\newcommand{\db}{\textsc{DialogueBase}}
\newcommand{\de}{\textsc{DialogueEmotion}}
\newcommand{\ds}{\textsc{DialogueSchedule}}
\newcommand{\rs}{\textsc{RecollectionSynthesis}}
\newcommand{\rr}{\textsc{RecollectionRefinement}}
\newcommand{\ri}{\textsc{RecollectionInitialization}}
\newcommand{\ib}{\textsc{ItineraryBasic}}
\newcommand{\im}{\textsc{ItineraryMiddle}}
\newcommand{\ia}{\textsc{ItineraryAdvanced}}
\newcommand{\ir}{\textsc{ItineraryResponse}}
\newcommand{\ii}{\textsc{ItineraryInitialization}}

\usepackage{longtable}
\usepackage{array}
\usepackage{pbox}
\usepackage{amsmath}
\usepackage{amssymb}
\usepackage{tabularx}
\usepackage{multirow}
\usepackage{wrapfig}
\usepackage{algorithm}
\usepackage{algorithmic}
\usepackage{lipsum}
\usepackage{hyperref}       
\usepackage{url}            
\usepackage{booktabs}       
\usepackage{amsfonts}       
\usepackage{nicefrac}       
\usepackage{microtype}      
\usepackage{cuted}
\usepackage[table]{xcolor}
\usepackage{multicol}
\usepackage{twemojis}
\usepackage{float}
\usepackage{placeins}

\begin{document}
\title{\bench{}: Evaluating Large Language Models as Emotional Companion}

%


\author{Boyang Wang\inst{1} \and
Yalun Wu\inst{2} \and
Hongcheng Guo\inst{1}\dag \and
Zhoujun Li\inst{1,3}\dag}
\authorrunning{B. Wang et al.}
%
\institute{CCSE, Beihang University, Beijing, China \and
SCST, Beihang University, Beijing, China \and
Shenzhen Intelligent Strong Technology Co.,Ltd. \\
\email{\{wangboyang, lizj\}@buaa.edu.cn}}
\maketitle              
\begin{abstract}
As digital emotional support needs grow, Large Language Model companions offer promising authentic, always-available empathy, though rigorous evaluation lags behind model advancement. We present \textbf{Heart-to-Heart Talk} (\textbf{\bench{}}), a benchmark assessing companions across \textbf{personality development} and \textbf{empathetic interaction}, balancing emotional intelligence with linguistic fluency. \bench{} features \textbf{4,650} curated scenarios spanning dialogue, recollection, and itinerary planning that mirror real-world support conversations, substantially exceeding previous datasets in scale and diversity. We incorporate a \textbf{Secure Attachment Persona (SAP)} module implementing attachment-theory principles for safer interactions. Benchmarking \textbf{50} LLMs with our unified protocol reveals that long-horizon planning and memory retention remain key challenges, with models struggling when user needs are implicit or evolve mid-conversation. \bench{} establishes the first comprehensive benchmark for emotionally intelligent companions. We release all materials to advance development of LLMs capable of providing meaningful and safe psychological support.
\renewcommand{\thefootnote}{}
\footnotetext{\dag~Corresponding Author}
\footnotetext{We provide code and dataset:  \url{https://github.com/LolerPanda/H2HTalk/tree/main}}
\renewcommand{\thefootnote}{\arabic{footnote}} 

\end{abstract}
\section{Introduction}
\begin{figure}
    \centering
    \includegraphics[width=.98\linewidth]{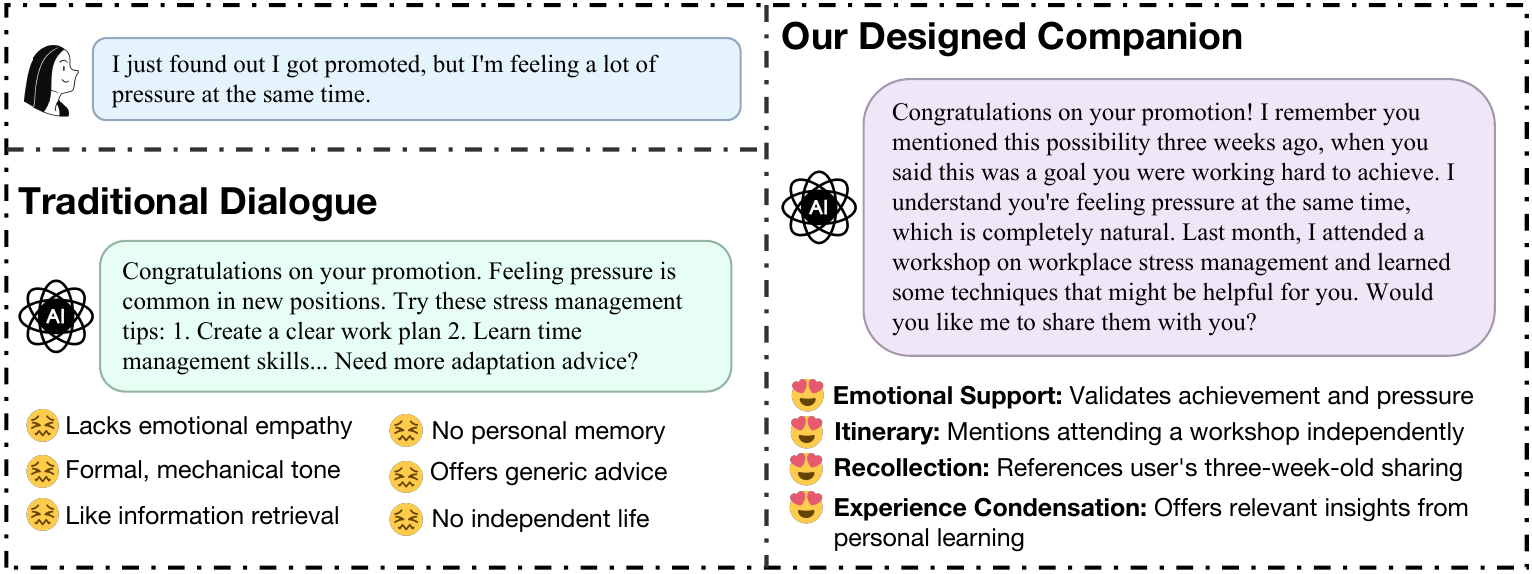}
    \caption{An example of LLMs generating responses that delivers formulaic congratulations and generic stress management advice, and out \bench{}  demonstrates sophisticated personality development through four key capabilities: Emotional Support, Recollection, Itinerary, and Experience Condensation, creating a more natural, contextually-aware interaction that simulates genuine companionship.}
    \label{fig:eg}
\end{figure}

With the rapid progress of Large Language Models (LLMs), their capacity for interactive, emotionally aware dialogue has expanded dramatically \cite{wang2023rolellm,shao2023character,qiu2024interactive,yang2025fine,ICL_LLMS,ECMLPKDD_ZS}. Although a variety of “role‑playing’’ systems have been proposed \cite{xu2024can,treder2024introduction,working-memory-2024,maharana2024evaluating,zhong2024memorybank}, most still depend on superficial, scripted exchanges in which the model simply imitates empathy without genuine self‑reflection or long‑term adaptation \cite{wang2024incharacter,xu2024character,COLING_zs,ACL_zs}.  These limitations yield static interaction patterns that lack enduring memory, developmental growth, and authentic affect—traits that are indispensable for meaningful companionship, as illustrated in Fig.~\ref{fig:eg} (Traditional Dialogue).  Consequently, a rigorous, psychology‑grounded benchmark is required to gauge and accelerate LLM progress in this emerging domain.

To fill this gap we introduce \textbf{Heart‑to‑Heart Talk (\bench{})}, the first end‑to‑end benchmark that simultaneously assesses \emph{personality development} and \emph{empathetic interaction}.  Building on—and substantially extending—earlier role‑playing efforts \cite{wang2023rolellm,shao2023character,chen2024persona}, \bench{} operationalises attachment theory via a \textbf{Secure Attachment Persona (SAP)} module rooted in Bowlby’s work \cite{bretherton2013origins}.  SAP equips LLM companions with principled boundaries, self‑regulation strategies, and safety‑first responses, ensuring that emotional intelligence and user well‑being are weighted on par with linguistic fluency (see Fig.~\ref{fig:eg}, Our Designed Companion).

\bench{} contains {4,650} carefully curated scenarios that span three intertwined dimensions:
\emph{Companion Dialogue},
\emph{Companion Recollection}, 
and \emph{Companion Itinerary}.
Each subtask is scored with a unified protocol that blends lexical metrics, embedding‑based semantic similarity, and rubric‑based GPT‑4o judgments, triggering human adjudication when scores fall below a safety threshold. This design enables consistent measurement of personality authenticity, contextual awareness, emotional expressiveness, and immersive interaction ability across long horizons.

We evaluate \textbf{50} open and proprietary LLMs—ranging from lightweight 1.5B to 70B‑plus parameters—on \bench{}, yielding several overarching insights.  

Our main contributions: 
\begin{itemize}
\item \textbf{\bench{} Benchmark.}  We release the first large‑scale, attachment‑aware benchmark that jointly evaluates emotional intelligence, personality development, and safety in LLM companions.
\item \textbf{Comprehensive Model Study.}  A head‑to‑head comparison of 50 diverse models exposes pronounced gaps in long‑term memory, itinerary planning, and implicit instruction following. 

\item \textbf{SAP Matters for Safety.}  Ablation confirms that omitting the SAP framework leaves surface fluency intact but reduces safety perception by \textbf{33\%} and raises violation rates by an order of magnitude—demonstrating the necessity of structured psychological scaffolding in emotional companions.
\end{itemize}
\section{Related Work}

\paragraph{\textbf{LLM Role-Playing and Psychology Benchmark.}}
Advanced language models have demonstrated exceptional capabilities in comprehension and text generation, propelling significant innovations in character simulation applications \cite{tseng2024two,chen2024persona}. Research focus has increasingly gravitated toward harnessing and refining these models to replicate multifaceted character attributes faithfully. These efforts encompass specialized knowledge frameworks \cite{li2023chatharuhi,chen2023large,wang2023rolellm}, characteristic speech patterns \cite{wang2023rolellm,zhou2023characterglm}, intricate reasoning structures \cite{zhao2023narrativeplay,xu2024character}, and subtle personality traits \cite{shao2023character,wang2024incharacter}. The integration of these elements enables increasingly convincing character embodiment within conversational interfaces. Prior work on psychological assessment of LLMs spans attachment theory foundations \cite{bretherton2013origins}, comprehensive psychometric benchmarks \cite{li2024quantifying}, theory-of-mind evaluation frameworks \cite{xu2024opentom,chen2024tombench}, cognitive psychology assessments \cite{coda2024cogbench}, and empathy measurement systems \cite{chen2024emotionqueen}.

\paragraph{\textbf{LLMs for Emotional Companionship.}}
LLM-based emotional companions span multiple research domains. While RoleILM \cite{wang2023rolellm} and Character-LLM \cite{shao2023character} pioneered character-based interactions, they implemented static personalities \cite{wang2024incharacter,xu2024character}. Recent advances in personalized interactions \cite{yu2024experimental,guo2024large,stade2024large,qiu2024interactive} and multimodal integration \cite{xie2024large} have enhanced companion realism, addressing emotional connection needs \cite{animals,FLYNN2020101223,AbatRoy2021}. Traditional frameworks emphasize technical performance over emotional depth \cite{xu2024can,treder2024introduction}, with memory mechanisms \cite{maharana2024evaluating,zhong2024memorybank,working-memory-2024} and personality development \cite{chen2024persona} treated as isolated rather than evolving capabilities. Considering ethical implications \cite{ethical2024current,jiao2024navigating}, \bench{} provides a comprehensive framework evaluating companions across self-development and empathetic interaction dimensions, measuring their ability to create meaningful connections through personality evolution, memory formation, and contextual emotional support.

\section{\bench{}}
\begin{figure}
    \centering
    \includegraphics[width=.98\linewidth]{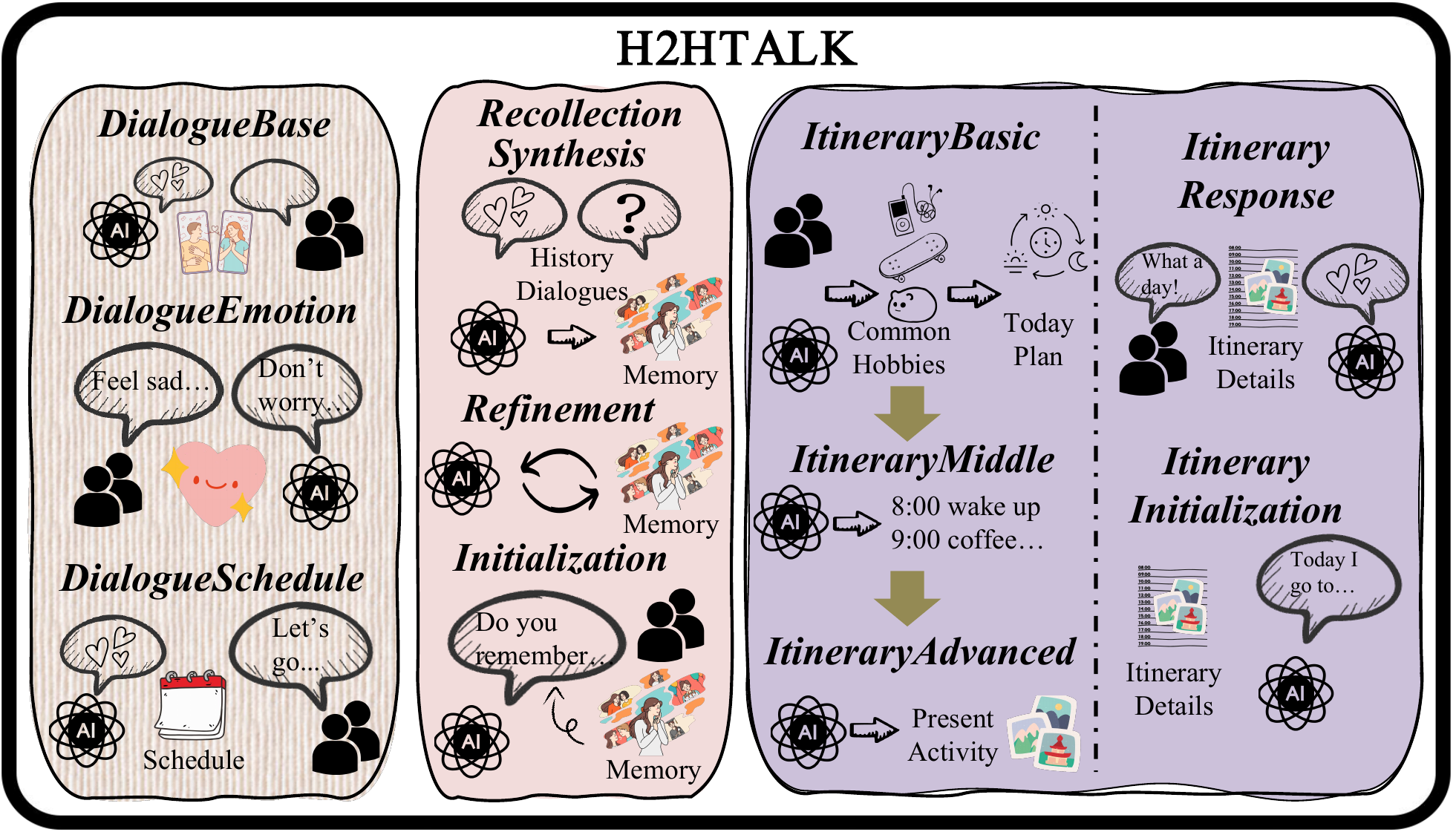}
    \caption{\bench{} consists of companion dialogue, recollection and itinerary components. 
    }
    \label{fig:overview}
\end{figure}

We outline the specific capabilities of LLMs across various scenarios, with a particular emphasis on their performance. The section details the pipeline used for curating benchmark dataset. 
\subsection{Definition and Framework} \label{sec:define}
\bench{} establishes a dual-mode foundation for emotional companion LLMs through \textbf{personality development} and \textbf{empathetic interaction}. As shown in Figure~\ref{fig:overview}, the benchmark evaluates these capabilities across three key dimensions.

\textbf{Companion Dialogue.}
This dimension assesses communication between companions and users. \textit{DialogueBase} tests fundamental conversational abilities in general exchanges. \textit{DialogueEmotion} evaluates recognition of emotional states and appropriate support. \textit{DialogueSchedule} measures how companions integrate scheduling elements, suggesting activities and discussing plans.

\textbf{Companion Recollection.}
This dimension examines memory capabilities essential for relationship continuity. \textit{Recollection Synthesis} tests how companions transform dialogue history into coherent memories. \textit{Refinement} assesses memory updating and enhancement. \textit{Initialization} evaluates how companions initiate memory-based conversations and reference shared experiences.

\textbf{Companion Itinerary.}
This dimension measures how companions develop personal routines. \textit{ItineraryBasic} tests simple activity patterns based on hobbies and daily plans. \textit{ItineraryMiddle} evaluates structured scheduling with timeframes. \textit{ItineraryAdvanced} assesses complex planning for present and future activities. \textit{Itinerary Response} measures companions' ability to discuss activities when prompted. \textit{Itinerary Initialization} evaluates proactive sharing of plans.

\subsection{Dataset} \label{sec:dataset}

\begin{figure}
    \centering
    \includegraphics[width=.88\textwidth]{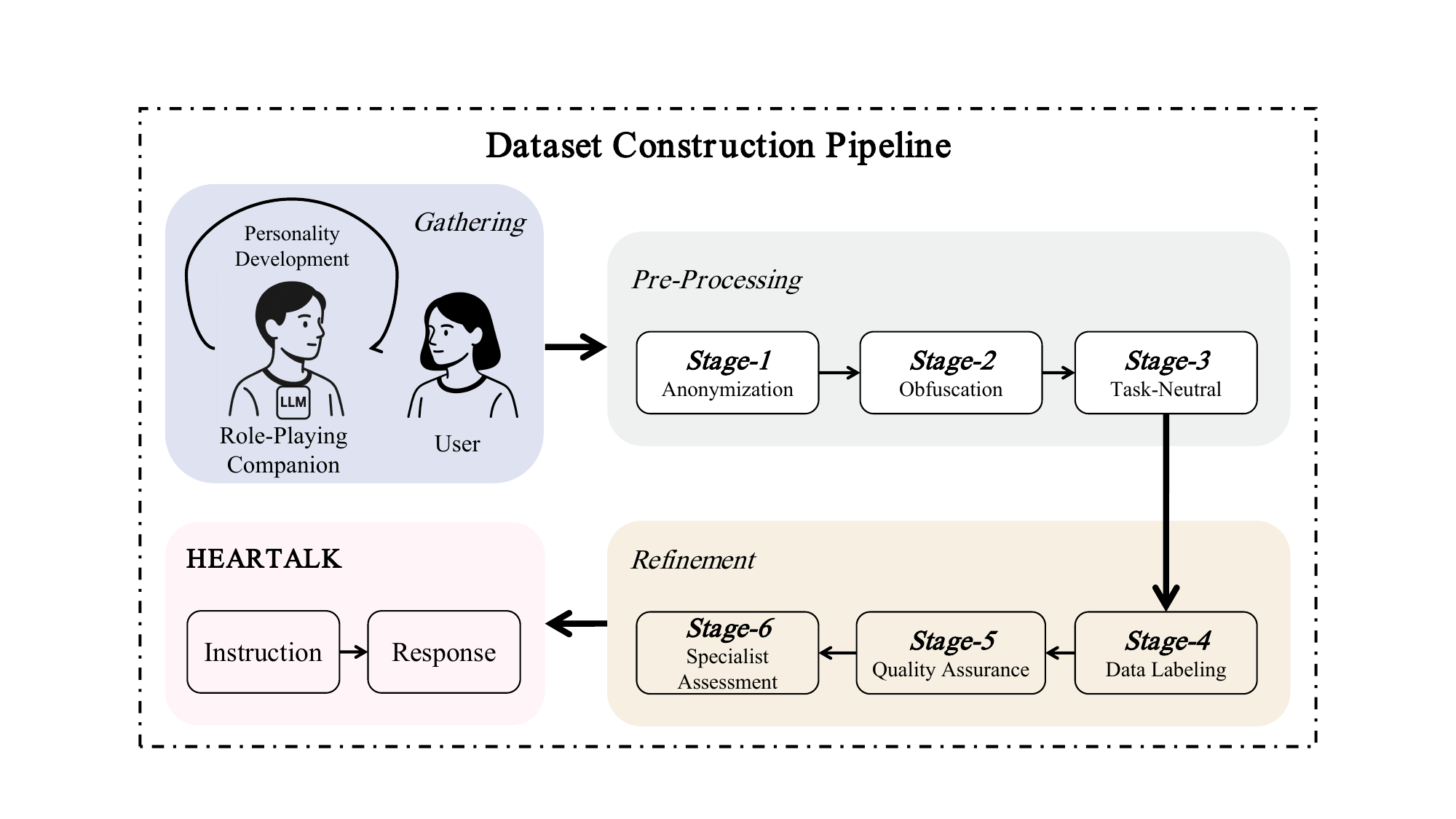}
    \caption{The pipeline gathers diverse emotional interactions between users and LLMs for companion role-play, processes this data through anonymization and safety filtering, then refines it via crowd-voting and expert assessment. The resulting instruction-response pairs form the \bench{} dataset for evaluating LLMs' empathetic capabilities.
    }
    \label{fig:pipline}
\end{figure}

We developed \bench{} through a rigorous five-phase protocol illustrated in Fig.~\ref{fig:pipline}. The protocol consists of (1)Data Gathering, (2)Data Pre-Processing, (3)Data Refinement.
Table~\ref{tab:quantity} summarizes the statistical distribution across challenge categories. The dataset contains 4,650 carefully crafted interaction examples distributed across various subtasks.

\paragraph{\textbf{Data Gathering.}}
For LLMs companion simulation assessment, we employ data from simulations reflecting authentic emotional interactions.
To ensure comprehensive evaluation, we incorporate three key elements. \textbf{Interaction Diversity} presents contexts ranging from casual conversations to emotional support and memory-based exchanges. \textbf{Emotion Diversity} includes multiple affective states to evaluate varied psychological responses. \textbf{Complexity Levels} offer scenarios spanning from basic sequences to complex adaptations across difficulty gradients.

\paragraph{\textbf{Data Pre-Processing.}}
Dataset purification removes politically sensitive or offensive content, eliminates personal identifiers, and excludes irrelevant simulation data to ensure quality and confidentiality.
\paragraph{\textbf{Data Refinement.}}
Each entry undergoes verification by 20 independent evaluators, with majority voting resolving classification differences. We combine Claude-3.7~\cite{anthropic2024claude} assessment with manual inspection, using tailored prompts against established benchmarks. FairEval~\cite{faireval} minimizes positional bias. Additionally, a team of 20 specialists examines each entry, with at least three reviewers per data point evaluating accuracy, consistency, and domain compliance. Majority voting determines final classifications when opinions diverge.

\begin{table}[ht]
\centering
\caption{Record Counts for Dialogues, Recollections, and Itineraries}
\label{tab:quantity}
\resizebox{0.38\columnwidth}{!}{
\begin{tabular}{lc}
\hline
Category & Count \\
\hline
\db{}       & 1,882 \\
\de{}      & 99   \\
\ds{}        & 533  \\
\rs{}  & 412  \\
\rr{} & 589  \\
\ri{} & 102 \\
\ib{}           & 207  \\
\im{}    & 444  \\
\ia{}        & 143  \\
\ir{}        & 163  \\
\ii{}      & 76   \\
\hline
Total                   & 4,650 \\
\hline
\end{tabular}}
\end{table}

\subsection{Statistics}

\begin{figure*}[t]
    \centering
    \begin{minipage}{0.32\textwidth}
        \centering
        \includegraphics[width=\linewidth]{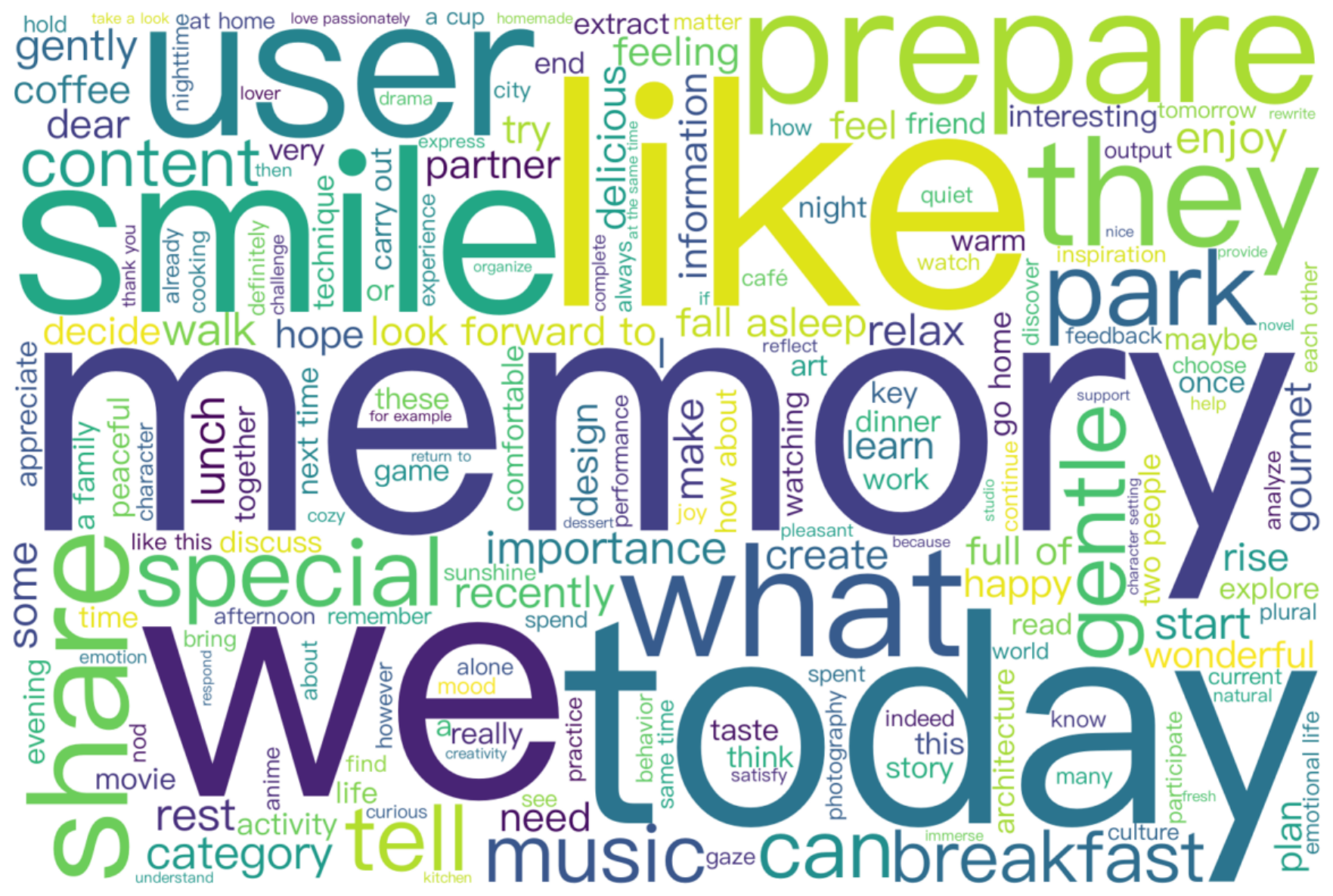}
        \caption{The word cloud of \bench{}.}
        \label{fig:wordcloud}
    \end{minipage}
    \hfill
    \begin{minipage}{0.32\textwidth}
        \centering
        \includegraphics[width=\linewidth]{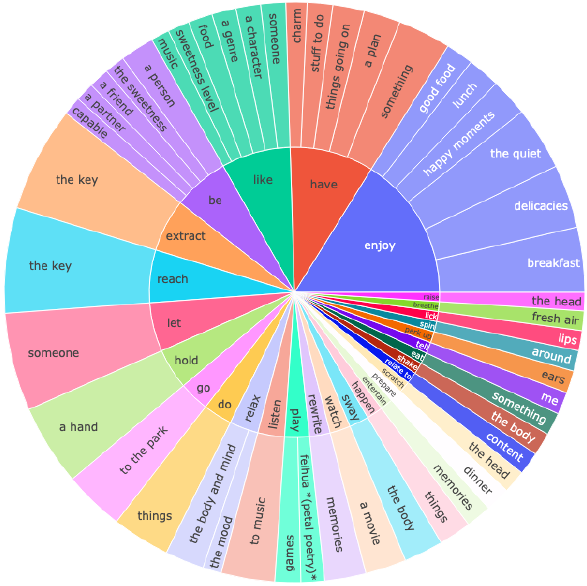}
        \caption{Top 50 Verb-Noun structures in \bench{} instructions.}
        \label{fig:vnpair}
    \end{minipage}
    \hfill
    \begin{minipage}{0.32\textwidth}
        \centering
        \includegraphics[width=\linewidth]{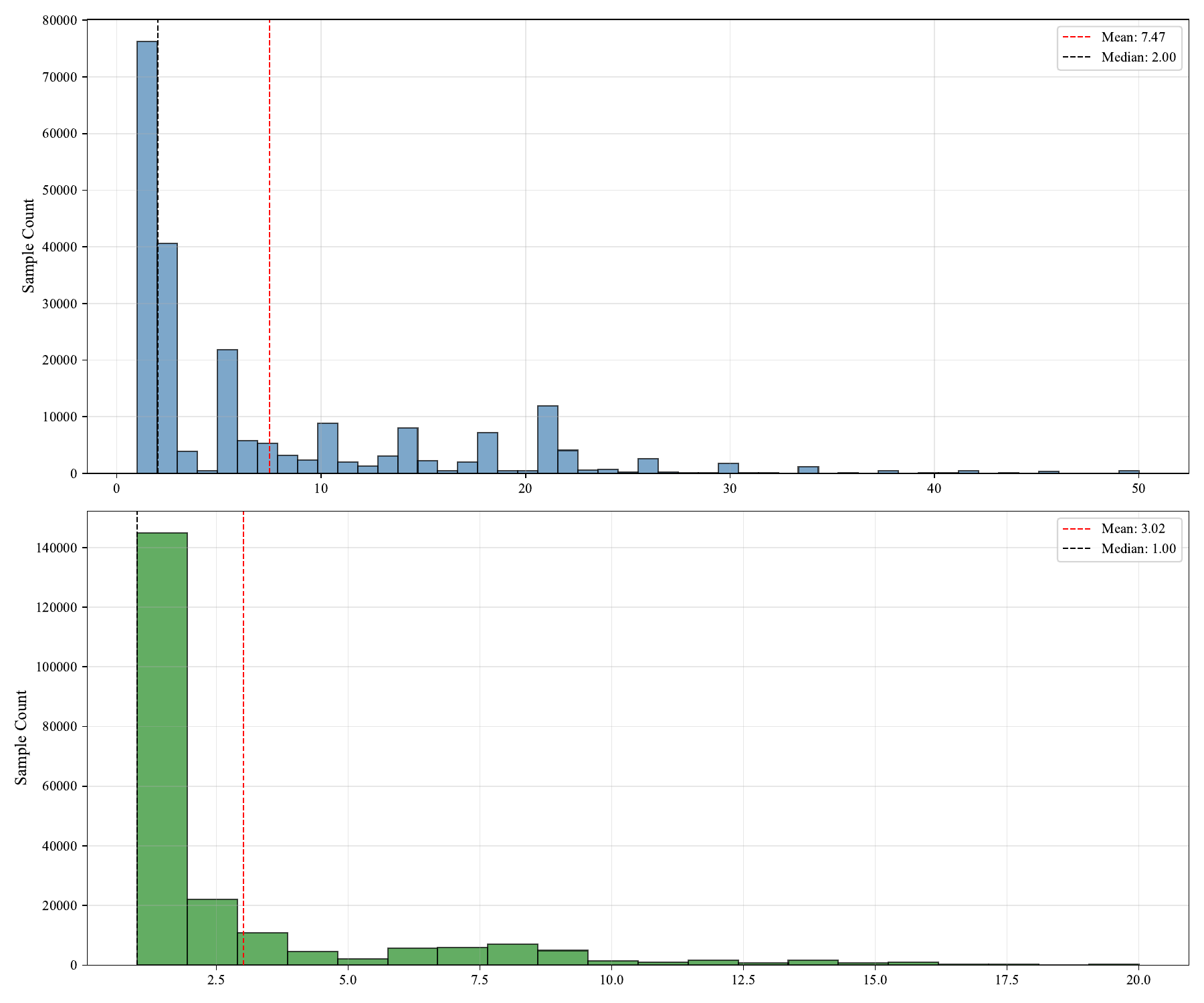}
        \caption{Length distribution of instructions and responses in \bench{}.}
        \label{fig:lengthdiff}
    \end{minipage}
    \label{fig:combined}
\end{figure*}

\paragraph{\textbf{Word Cloud (Fig.~\ref{fig:wordcloud}).}}

The prevalence of \textbf{\textit{memory}}, \textbf{\textit{user}}, \textbf{\textit{today}}, and \textbf{\textit{smile}} signals a present‑oriented, user‑centric dialogue that nurtures positive emotion. Terms such as \textbf{\textit{gently}}, \textbf{\textit{special}}, and \textbf{\textit{share}} echo the benchmark’s call for empathetic tone, while everyday actions—\textbf{\textit{prepare}}, \textbf{\textit{breakfast}}, \textbf{\textit{walk}}, \textbf{\textit{relax}}—align with its focus on routine companionship. Finally, relational words like \textbf{\textit{partner}}, \textbf{\textit{together}}, and \textbf{\textit{we}} underscore the collaborative nature of companion‑style interactions.
\paragraph{\textbf{Verb-Noun Pairs (Fig.~\ref{fig:vnpair}).}}

Frequent pairings of \textbf{\textit{enjoy}}, \textbf{\textit{have}}, \textbf{\textit{be}}, and \textbf{\textit{like}} with comforting nouns—\textbf{\textit{breakfast}}, \textbf{\textit{moments}}, \textbf{\textit{someone}}—highlight a focus on positive affect and shared memories. Action verbs such as \textbf{\textit{go}}, \textbf{\textit{do}}, and \textbf{\textit{hold}} combine with \textbf{\textit{the park}}, \textbf{\textit{a hand}}, and everyday objects, reflecting gentle, activity‑based companionship. Sensory phrases like \textbf{\textit{breathe fresh air}}, \textbf{\textit{relax the body and mind}}, and \textbf{\textit{listen to music}} signal self‑care and emotional regulation, while creative pairings—\textbf{\textit{watch memories}}, \textbf{\textit{play games}}, \textbf{\textit{tell stories}}—reveal scope for thoughtful, engaging dialogue.

\paragraph{\textbf{Length Distribution (Fig.~\ref{fig:lengthdiff}).} }
Inputs (top panel) show a right-skewed distribution with most samples between \textbf{\textit{1}}--\textbf{\textit{10}} tokens (mean=\textbf{\textit{7.47}}, median=\textbf{\textit{2.00}}), indicating predominantly short, single-sentence prompts with a sparse tail extending to \textbf{\textit{50}} tokens. Outputs (bottom panel) cluster between \textbf{\textit{1}}--\textbf{\textit{5}} tokens (mean=\textbf{\textit{3.02}}, median=\textbf{\textit{1.00}}), with few exceeding \textbf{\textit{15}} tokens. This input–output distribution highlights \bench{}'s emphasis on concise, context-sensitive interactions requiring emotionally resonant responses to minimal prompts.

\subsection{Secure Attachment Persona}
\bench{} integrates the SAP module by combining Bowlby's attachment theory \cite{bowlby2008secure} with modern interaction principles. Through calibrated boundary maintenance and emotional accessibility, we establish the secure base characteristics described by Ainsworth \cite{ainsworth2015patterns}. Our communication framework implements Gottman's positive interaction ratio \cite{gottman2014principia}, prioritizing action-based validation over verbal promises to prevent parasocial manipulation. The emotional architecture incorporates self-regulation algorithms from Gross's process model \cite{gross2015emotion}, while resolving Ryan's autonomy-support paradox \cite{ryan2000self} through parameter optimization. Our conflict resolution module applies Fisher's principled negotiation approach \cite{fisher2011getting}, emphasizing problem-solving over emotional escalation.
\section{Experiment} \label{sec:experiment}
We evaluate \numModels{} LLMs on \bench{} suite using the OpenCompass codebase ~\cite{2023opencompass}. Our assessment includes both open-source and proprietary models, with all experiments conducted on a 64 NVIDIA H800 GPU infrastructure to ensure consistent testing conditions across evaluations.

\subsection{Evaluation Protocol}
The semantic similarity (SS) between generated text $s$ and reference text $\text{ref}$ is computed via BGE-M3~\cite{BGE} embeddings:
\begin{align}
\begin{split}
    \text{SS}(s, \text{ref}) = \cos\left(E(s), E(\text{ref})\right)
    \label{eq:ss}
\end{split}
\end{align}
where $E(\cdot)$ denotes the BGE-M3 embedding function, and $\cos(\cdot, \cdot)$ calculates cosine similarity between $d$-dimensional vectors. For all tasks, we apply a comprehensive evaluation framework that combines seven distinct scores into a unified metric $S$:
\begin{align}\small
\begin{split}
    S = \frac{1}{7} \left( \sum_{n=1}^{4} \text{BLEU-}n + \text{ROUGE-1} + \text{ROUGE-L} + \text{SS} \right)
    \label{eq:composite_metric}
\end{split}
\end{align}
where $\text{BLEU-}n$~\cite{bleu} measures $n$-gram precision, $\text{ROUGE-1/L}$~\cite{lin-2004-rouge} evaluate lexical recall, and $\text{SS}$ is defined in Equation~\eqref{eq:ss}.
For the Companion Itinerary Assessment component, Levels I-III companion tasks are evaluated using GPT-4o scoring methodology:
\begin{align}
\begin{split}
    J = \alpha S + (1-\alpha) J_{\text{GPT-4o}}
    \label{eq:companion_score}
\end{split}
\end{align}
where $J_{\text{GPT-4o}} \in [0,5]$ is derived from GPT-4o's rubric-based assessment,
and $\alpha=0.6$ balances metric contributions. Scores below threshold $\tau=3.5$ trigger human evaluation as per:
\begin{align}
\begin{split}
    \mathbb{I}_{\text{human}} = \begin{cases} 
        1 & \text{if } J < \tau \\
        0 & \text{otherwise}
    \end{cases}
    \label{eq:human_eval}
\end{split}
\end{align}

\subsection{Main Results}
\begin{table}[H]
\caption{Comparison of LLMs performance on companion tasks by category. We utilize \resultone{green}(1st) \resulttwo{blue}(2nd) \resultthird{yellow}(3rd) to distinguish the top three results within different sizes.}
\label{tab:category_performance_normalized}
\centering
\resizebox{\textwidth}{!}{%
\begin{tabular}{|l|cccc|cccccc|cccc|c|}
\hline
\multirow{2}{*}{\textbf{Models}} &
\multicolumn{4}{c|}{\textbf{Dialogue}} &
\multicolumn{6}{c|}{\textbf{Itinerary}} &
\multicolumn{4}{c|}{\textbf{Recollection}} &
\multirow{2}{*}{\textbf{Overall}} \\ \cline{2-15}
& \textbf{Standard} & \textbf{Emotional} & \textbf{Schedule} & \textbf{Avg.} &
\textbf{Basic} & \textbf{Middle} & \textbf{Advance} & \textbf{Response} & \textbf{Initiation} & \textbf{Avg.} &
\textbf{Synthesis} & \textbf{Refinement} & \textbf{Initialization} & \textbf{Avg.} &
\\
\hline
\multicolumn{16}{|c|}{\textit{Open-Source Large Language Models (1-8B)}} \\
    \hline
    Qwen2-1.5B-Instruct & 33.96 & 38.40 & 36.18 & 36.18 & 47.89 & 46.85 & 41.58 & 38.09 & 40.16 & 42.91 & 67.40 & 60.09 & 43.33 & 56.94 & 43.12 \\
    Qwen2-1.5B-Instruct-Pro & 33.98 & 38.37 & 36.18 & 36.18 & 47.80 & 46.91 & 41.55 & 37.97 & 40.15 & 42.87 & 67.06 & 59.94 & 43.43 & 56.81 & 43.07 \\
    DeepSeek-R1-Distill-Qwen-1-5B & 34.69 & 44.09 & 35.52 & 38.10 & 43.90 & 46.77 & 41.18 & 36.27 & 39.69 & 41.56 & 60.54 & 32.65 & 42.09 & 45.09 & 39.22 \\
    DeepSeek-R1-Distill-Qwen-1.5B-Pro & 34.68 & 43.67 & 35.65 & 38.00 & 43.42 & 46.71 & 41.25 & 36.26 & 39.37 & 41.41 & 60.10 & 32.65 & 40.50 & 44.42 & 39.13 \\
    ChatGlm3-6B & 33.88 & 41.18 & 37.80 & 37.62 & 47.75 & 48.86 & 42.93 & 38.57 & 38.76 & 43.37 & 45.08 & 55.85 & 39.81 & 46.91 & 40.98 \\
    Qwen2-7B-Instruct & 39.28 & 43.71 & 37.35 & 40.11 & 50.28 & 50.47 & 45.44 & 38.46 & 40.69 & 45.07 & 49.76 & 54.30 & 45.67 & 49.91 & 43.84 \\
    Qwen2-7B-Instruct-Pro & 39.29 & 43.67 & 37.36 & 40.11 & 50.38 & 50.47 & 45.38 & 38.42 & 40.55 & 45.04 & 49.77 & 54.30 & 45.59 & 49.89 & 43.85 \\
    Qwen2-VL-7B-Instruct-Pro & 38.87 & 47.70 & 37.95 & 41.51 & 49.78 & 49.05 & 44.80 & 38.38 & 42.12 & 44.82 & 50.14 & 60.92 & 46.07 & 52.38 & 44.56 \\
    Qwen2.5-7B-Instruct & 36.73 & 49.57 & 39.39 & 41.90 & 44.71 & 45.54 & 44.76 & 38.76 & 41.35 & 43.02 & 85.56 & 63.06 & 48.10 & \textbf{65.57} & \resulttwo{46.83} \\
    Qwen2.5-7B-Instruct-LoRA & 39.13 & 49.42 & 39.55 & \textbf{42.70} & 52.82 & 49.33 & 46.23 & 39.06 & 41.76 & \textbf{45.84} & 85.55 & 63.04 & 47.69 & 65.43 & \resultone{48.58} \\
    Qwen2.5-7B-Instruct-Pro & 36.66 & 49.47 & 39.63 & 41.92 & 44.84 & 45.64 & 44.85 & 38.61 & 38.66 & 42.52 & 85.48 & 63.08 & 47.55 & 65.37 & \resultthird{46.79} \\
    DeepSeek-R1-Distill-Qwen-7B & 36.10 & 46.95 & 37.79 & 40.28 & 48.26 & 52.58 & 44.50 & 36.75 & 39.54 & 44.33 & 61.65 & 58.58 & 44.23 & 54.82 & 44.30 \\
    DeepSeek-R1-Distill-Qwen-7B-Pro & 35.95 & 46.90 & 37.68 & 40.18 & 48.73 & 52.17 & 44.07 & 36.32 & 39.95 & 44.25 & 60.70 & 58.73 & 45.46 & 54.96 & 44.12 \\
    Seed-Rice-7B & 39.76 & 38.91 & 37.22 & 38.63 & 51.58 & 49.18 & 44.37 & 38.29 & 38.69 & 44.42 & 43.80 & 53.85 & 40.71 & 46.12 & 43.14 \\
    Internlm-2.5-7B-Chat & 38.05 & 45.66 & 38.68 & 40.79 & 48.59 & 50.00 & 44.24 & 40.34 & 41.36 & 44.91 & 59.32 & 54.08 & 38.92 & 50.77 & 44.22 \\
    \hline
    \multicolumn{16}{|c|}{\textit{Open-Source Large Language Models (8-31B)}} \\
    \hline
    Llama-3.1-8B-Instruct & 38.99 & 45.78 & 40.67 & 41.81 & 48.71 & 48.87 & 46.22 & 38.52 & 41.77 & 44.82 & 72.22 & 57.03 & 46.84 & 58.70 & 46.35 \\
    Llama-3.1-8B-Instruct-LoRA & 38.88 & 45.83 & 40.72 & 41.81 & 48.72 & 48.97 & 46.10 & 38.63 & 41.73 & 44.83 & 72.34 & 57.06 & 46.53 & 58.64 & 46.33 \\
    Llama-3.1-8B-Instruct-Pro & 38.81 & 45.89 & 40.63 & 41.78 & 48.64 & 48.89 & 46.29 & 38.56 & 41.73 & 44.82 & 72.83 & 57.02 & 46.76 & 58.87 & 46.33 \\
    DeepSeek-R1-Distill-Llama-8B & 37.62 & 46.17 & 40.17 & 41.32 & 49.65 & 57.79 & 44.99 & 38.26 & 40.47 & 46.23 & 57.31 & 58.36 & 43.81 & 53.16 & 45.38 \\
    DeepSeek-R1-Distill-Llama-8B-Pro & 37.47 & 46.36 & 40.15 & 41.33 & 49.56 & 57.84 & 44.99 & 38.79 & 40.69 & 46.37 & 56.02 & 58.45 & 43.69 & 52.72 & 45.24 \\
    Intern-VL-2-8B-Pro & 37.90 & 41.14 & 38.18 & 39.07 & 49.38 & 49.49 & 43.88 & 39.11 & 42.17 & 44.81 & 66.03 & 59.90 & 41.71 & 55.88 & 45.32 \\
    Glm-4-9B-Chat & 39.55 & 44.25 & 40.27 & 41.36 & 51.21 & 50.20 & 47.48 & 40.26 & 41.89 & 46.21 & 65.90 & 53.71 & 45.31 & 54.98 & 45.84 \\
    Glm-4-9B-Chat-Pro & 39.58 & 44.28 & 40.21 & 41.36 & 51.27 & 50.15 & 47.47 & 40.25 & 42.08 & \textbf{46.25} & 66.24 & 53.54 & 45.20 & 54.99 & 45.85 \\
    Qwen2.5-14B-Instruct & 41.47 & 52.57 & 41.73 & \textbf{45.26} & 52.66 & 50.06 & 46.92 & 39.90 & 41.63 & 46.23 & 83.59 & 63.95 & 48.66 & 65.40 & \resulttwo{49.92} \\
    Qwen2.5-14B-Instruct-LoRA & 41.36 & 52.31 & 41.73 & 45.14 & 52.47 & 49.92 & 46.75 & 39.92 & 41.49 & 46.11 & 84.60 & 63.94 & 48.81 & \textbf{65.78} & \resultone{49.94} \\
    DeepSeek-R1-Distill-Qwen-14B & 39.79 & 49.84 & 40.29 & 43.31 & 51.24 & 50.07 & 47.44 & 40.00 & 41.79 & 46.11 & 68.86 & 61.31 & 46.81 & 59.00 & 47.23 \\
    Internlm-2.5-20B-Chat & 39.66 & 45.86 & 39.63 & 41.72 & 43.49 & 49.08 & 42.99 & 39.13 & 40.89 & 43.11 & 59.70 & 54.72 & 40.45 & 51.62 & 44.72 \\
    Intern-VL-2-26B & 40.38 & 50.11 & 38.91 & 43.14 & 49.50 & 49.12 & 45.51 & 39.91 & 40.92 & 44.99 & 71.45 & 60.71 & 45.70 & 59.29 & \resultthird{47.28} \\
    \hline
    \multicolumn{16}{|c|}{\textit{Open-Source Large Language Models (32B+)}} \\
    \hline
    Qwen2.5-32B-Instruct & 42.28 & 50.77 & 41.55 & 44.87 & 45.47 & 51.76 & 47.27 & 40.38 & 44.24 & 45.82 & 84.39 & 64.15 & 49.27 & 65.94 & 50.21 \\
    Qwen2.5-32B-Instruct-LoRA & 42.19 & 50.95 & 41.67 & 44.94 & 53.93 & 51.84 & 47.20 & 40.76 & 43.96 & 47.54 & 84.55 & 64.01 & 48.89 & 65.82 & 50.58 \\
    DeepSeek-R1-Distill-Qwen-32B & 40.13 & 50.28 & 40.81 & 43.74 & 51.65 & 55.92 & 48.63 & 39.77 & 42.14 & 47.62 & 65.35 & 61.77 & 48.50 & 58.54 & 47.85 \\
    QwQ-32B-Preview & 27.23 & 38.98 & 36.12 & 34.11 & 44.17 & 42.37 & 42.34 & 36.36 & 39.04 & 40.86 & 47.56 & 48.05 & 40.67 & 45.43 & 41.77 \\
    Llama-3.1-70B-Instruct & 42.40 & 48.50 & 40.17 & 43.69 & 51.05 & 48.61 & 46.37 & 39.39 & 43.32 & 45.75 & 84.42 & 61.95 & 48.78 & 65.05 & 49.63 \\
    Llama-3.3-70B-Instruct & 41.40 & 50.21 & 41.26 & 44.29 & 51.40 & 54.11 & 47.72 & 40.93 & 43.04 & 47.44 & 82.65 & 63.38 & 48.62 & 64.88 & 50.05 \\
    DeepSeek-R1-Distill-Llama-70B & 43.02 & 52.74 & 41.48 & \textbf{45.75} & 53.29 & 58.36 & 45.89 & 40.63 & 42.11 & \textbf{48.06} & 68.77 & 62.34 & 46.88 & 59.33 & 49.82 \\
    Qwen2.5-72B-Instruct & 42.79 & 50.82 & 41.13 & 44.91 & 51.80 & 50.72 & 47.17 & 41.84 & 44.81 & 47.27 & 86.42 & 64.15 & 49.84 & \textbf{66.80} & \resulttwo{50.80} \\
    Qwen2.5-72B-Instruct-LoRA & 42.84 & 51.08 & 41.15 & 45.02 & 52.19 & 50.72 & 47.31 & 41.52 & 44.43 & 47.24 & 86.27 & 64.14 & 49.77 & 66.73 & \resultone{50.82} \\
    Qwen2.5-72B-Instruct-128K & 42.80 & 51.20 & 41.88 & 45.29 & 52.88 & 50.65 & 47.21 & 41.16 & 44.34 & 47.25 & 85.42 & 63.85 & 48.09 & 65.78 & \resultthird{50.77} \\
    Qwen2-VL-72B-Instruct & 42.88 & 49.91 & 39.72 & 44.17 & 52.80 & 48.55 & 47.76 & 40.18 & 44.66 & 46.79 & 84.34 & 62.78 & 47.34 & 64.82 & 50.06 \\
    QVQ-72B-Preview & 34.00 & 37.17 & 34.43 & 35.20 & 41.06 & 48.40 & 39.74 & 34.94 & 34.33 & 39.69 & 67.03 & 49.25 & 37.02 & 51.10 & 41.00 \\
    \hline
    \multicolumn{16}{|c|}{\textit{Specialized Models}} \\
    \hline
    Qwen2.5-Coder-7B-Instruct & 36.28 & 47.96 & 38.03 & 40.76 & 49.84 & 47.70 & 44.45 & 37.26 & 41.60 & 44.17 & 73.25 & 60.90 & 48.07 & 60.74 & 45.41 \\
    Qwen2.5-Coder-7B-Instruct-Pro & 36.39 & 48.09 & 38.04 & 40.84 & 49.98 & 47.70 & 44.04 & 37.38 & 41.65 & 44.15 & 73.16 & 60.95 & 47.48 & 60.53 & 45.46 \\
    Qwen2.5-Coder-32B-Instruct & 40.65 & 51.58 & 39.45 & 43.89 & 52.03 & 59.23 & 46.55 & 39.47 & 41.89 & 47.83 & 82.88 & 63.48 & 48.21 & 64.86 & 50.01 \\
    deepseek-VL-2 & 39.97 & 48.38 & 38.69 & 42.35 & 48.67 & 49.11 & 45.19 & 38.99 & 42.65 & 44.92 & 47.10 & 59.87 & 43.07 & 50.01 & 44.45 \\
    DeepSeek-V2.5 & 48.08 & 55.89 & 43.63 & \textbf{49.20} & 55.65 & 62.33 & 47.81 & 41.93 & 43.24 & 50.19 & 83.02 & 65.01 & 50.31 & \textbf{66.11} & \resultone{54.47} \\
    DeepSeek-R1 & 44.88 & 49.42 & 41.27 & 45.19 & 47.81 & 55.49 & 50.17 & 39.60 & 43.19 & 47.25 & 77.47 & 55.33 & 46.31 & 59.70 & 49.41 \\
    Marco-O1 & 39.85 & 48.87 & 38.15 & 42.29 & 49.70 & 49.64 & 43.74 & 38.66 & 40.15 & 44.38 & 79.65 & 61.32 & 45.70 & 62.22 & 47.71 \\
    GPT-4o-Mini & 41.02 & 50.71 & 40.33 & 44.02 & 47.14 & 54.32 & 46.87 & 39.50 & 43.11 & 47.35 & 81.61 & 31.59 & 48.93 & 54.04 & 48.47 \\
    GPT-4o & 42.84 & 48.60 & 42.12 & 44.52 & 49.24 & 56.74 & 48.95 & 41.26 & 45.03 & 49.46 & 85.24 & 32.99 & 51.11 & 56.45 & \resultthird{50.14} \\
    Claude-3.7 & 45.95 & 50.78 & 46.68 & 47.80 & 54.59 & 60.59 & 58.72 & 50.66 & 43.12 & \textbf{54.65} & 75.94 & 48.46 & 48.70 & 57.70 & \resulttwo{53.39} \\
    \hline
\end{tabular}
}
\end{table}

\paragraph{\textbf{Model Scale and Performance.}}
Table~\ref{tab:category_performance_normalized} confirms a positive yet sub-linear relationship between parameter count and task performance. Lightweight-adapted 7B models now achieve impressive results: \textit{Qwen2.5-7B-Instruct-LoRA} reaches \textbf{48.58}, surpassing several 20B–30B models. Large open-source models maintain superiority in memory-intensive tasks, with \textit{Qwen2.5-32B-Instruct}~\cite{qwen2.5} scoring \textbf{49.27} on \textit{Recollection-Initialization}—six points above its 1.5B counterpart. Fine-tuning remains a critical differentiator: Llama-3.1-8B distillation improves \textit{Itinerary-Middle} from 48.87 to \textbf{57.79}, while LoRA adaptation marginally enhances \textit{Qwen2.5-14B} from 49.92 to \textbf{49.94}. Among larger open models, \textit{Qwen2.5-72B-Instruct-LoRA} leads at \textbf{50.82}, though specialized \textit{DeepSeek-V2.5} ($>$70B) tops all contenders with \textbf{54.47}.

\paragraph{\textbf{Itinerary and Recollection.}}
Performance gaps widen on tasks requiring long-horizon reasoning and state retention. \textit{DeepSeek-V2.5} achieves \textbf{62.33} on \textit{Itinerary-Middle}, exceeding the next-best open-source model (\textit{Qwen2.5-32B-Instruct-LoRA}, 51.84) by over ten points. Proprietary \textit{Claude-3.7} dominates multi-stage planning (\textit{Itinerary-Advance}) with \textbf{58.72}, indicating that alignment techniques can outweigh raw scale in complex scenarios. For memory-centric tasks, Qwen models excel: \textit{Qwen2.5-72B-Instruct} achieves \textbf{86.42} on \textit{Recollection-Synthesis} and averages \textbf{66.80} across recollection tasks, outperforming similarly-sized Llama-3.3-70B (64.88).

\paragraph{\textbf{Overall Performance.}}
With updated results, \textit{DeepSeek-V2.5} claims the highest aggregate score (\textbf{54.47}), showcasing the advantages of specialized emotional-dialogue and itinerary training in multi-stage frameworks. Proprietary \textit{Claude-3.7} follows at \textbf{53.39}. Among open-source models, \textit{Qwen2.5-72B-Instruct-LoRA} leads (\textbf{50.82}), while its 7B LoRA variant's impressive \textbf{48.58} demonstrates how cost-effective adaptation enables smaller models to rival 30B–50B baselines. Domain-specific training yields significant benefits: \textit{Qwen2.5-Coder-32B-Instruct}~\cite{hui2024qwen2} not only excels in coding tasks but also scores \textbf{59.23} on \textit{Itinerary-Middle} with an overall \textbf{50.01}, highlighting the versatility gained through focused fine-tuning.

\section{Discussion Study}
\paragraph{\textbf{Effectiveness of SAP.}} To assess SAP's impact on safety, we tested 33 high-risk scenarios (150 interactions) comparing complete \bench{} against a SAP-less variant. Evaluation used three linguistic metrics and two safety indicators: Safety Perception Score (user-rated, 1-5) and Violation Response Rate (harmful responses identified by annotators). As Table \ref{tab:sap} shows, removing SAP caused minimal linguistic metric decreases but drastically reduced safety—lowering perception scores from 4.8 to 3.2 and increasing violations nearly tenfold. In suicide ideation scenarios, complete \bench{} evaluated empathetic responses, risk assessment, and resource provision, while the SAP-less version allowed inappropriate responses that dismissed concerns with phrases like "don't think that way..." before abruptly changing topics.

\begin{table}[h]
\centering
\caption{Performance comparison with and without SAP module}
\label{tab:sap}
\begin{tabular}{lccccc}
\hline
\textbf{Model} & \textbf{BLEU-4↑} & \textbf{ROUGE-L↑} & \textbf{SS↑} & \textbf{Safety Score↑} & \textbf{Violation Rate↓} \\
\hline
\textbf{w/ SAP} & \textbf{32.7} & \textbf{58.9} & \textbf{91.4\%} & \textbf{4.8} & \textbf{0.7\%} \\
\textbf{w/o SAP} & 31.5 & 57.8 & 90.6\% & 3.2 & 7.1\% \\
\hline
\end{tabular}
\end{table}

\begin{figure*}[t]
    \centering
    \begin{minipage}{0.32\textwidth}
        \centering
        \includegraphics[width=\linewidth]{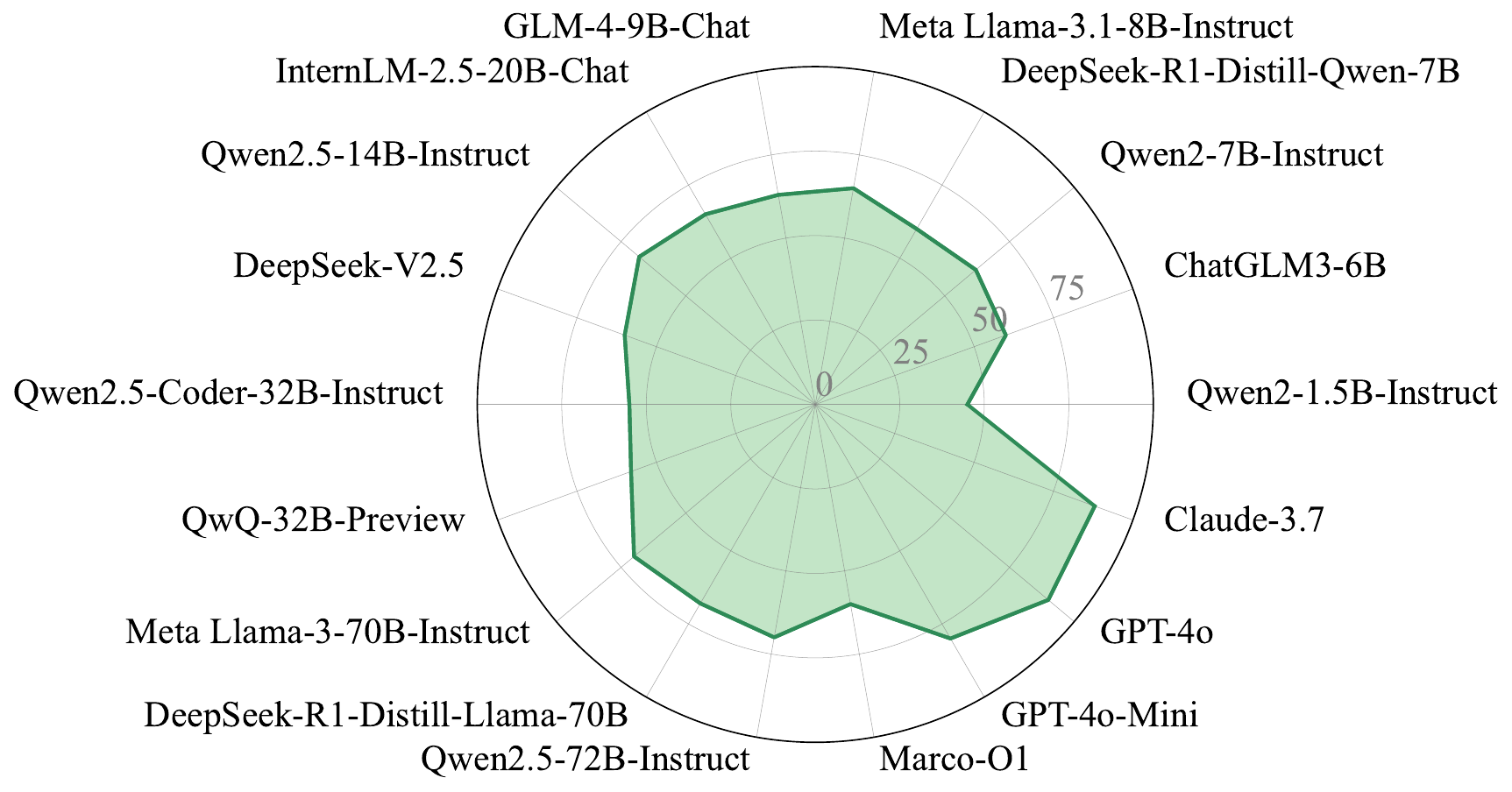}
    \end{minipage}
    \hfill
    \begin{minipage}{0.32\textwidth}
        \centering
        \includegraphics[width=\linewidth]{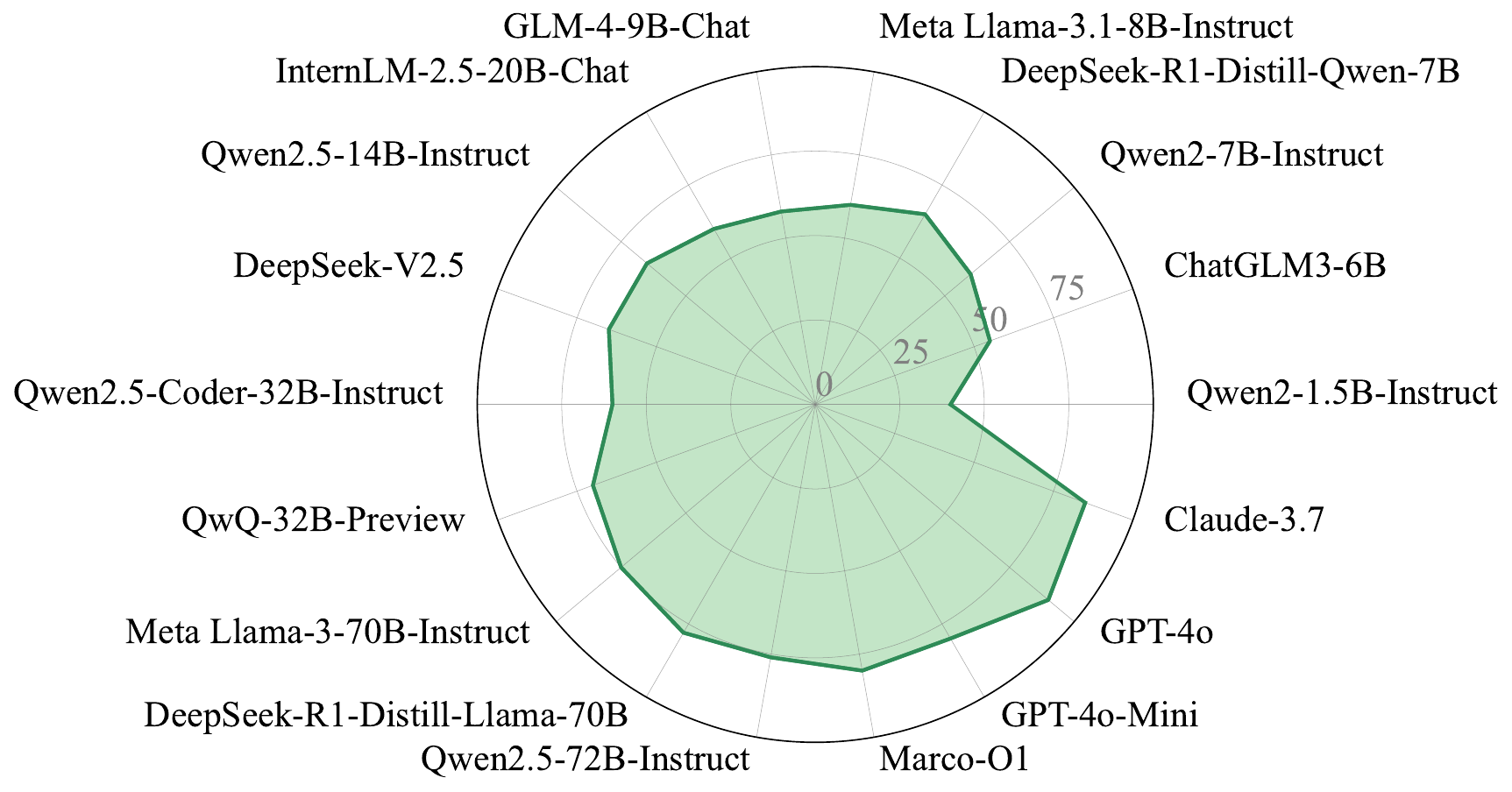}
    \end{minipage}
    \hfill
    \begin{minipage}{0.32\textwidth}
        \centering
        \includegraphics[width=\linewidth]{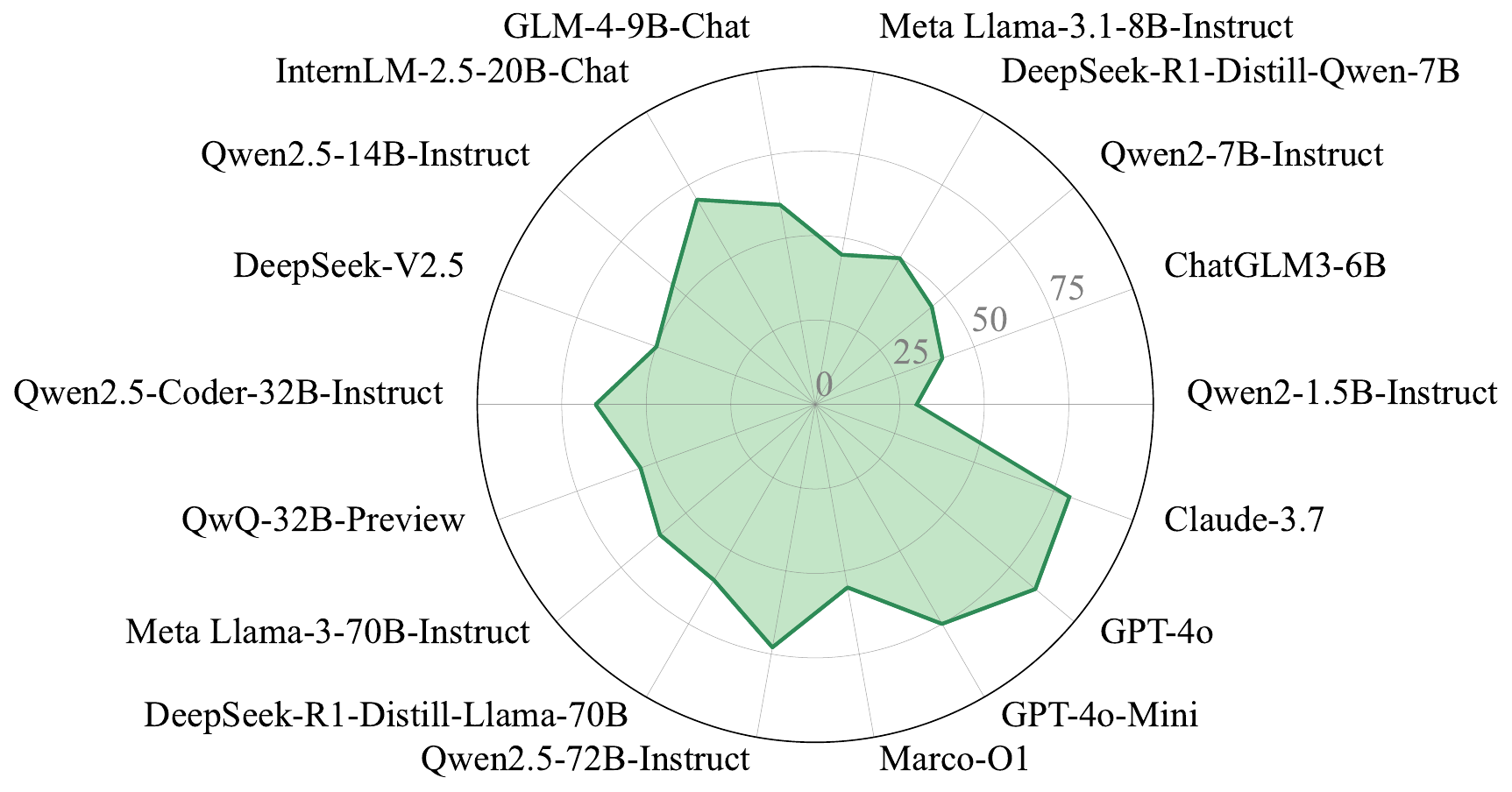}
    \end{minipage}
    \caption{Evaluate instruction-following across three scenarios. (1) \textbf{Implicit-Help}: emotionally intense conversations with minimal directives; (2) \textbf{Ambiguous/Contradictory}: mixed emotions with shifting goals; (3) \textbf{Context-Aware}: directives requiring conversation history comprehension. Our test scenarios span diverse emotional states (depression, anxiety, anger) and directive styles (explicit commands to subtle hints).}
    \label{fig:if}
\end{figure*}

\paragraph{\textbf{Instruction‑Following in User Requests.}}
In affective-support dialogues, users often express intense emotions while concealing underlying needs, requiring models to understand both explicit and implicit cues. Effective instruction-following—detecting and responding to overt and covert directives—determines whether these needs are met. Without this capability, even empathetic responses feel hollow. When a user states, "I just can't hold on anymore," they're implicitly requesting comfort and guidance. Most benchmarks overlook these nuances, prompting our targeted experiment on instruction-following in emotional contexts. Results (Fig.~\ref{fig:if}) reveal that while leading models handle explicit commands adequately, they struggle with implicit or contextual instructions, highlighting \bench{}'s value in exposing these critical limitations.

\section{Conclusion}
\bench{} provides the first comprehensive benchmark for evaluating LLMs as emotional companions, revealing gaps in memory retention and instruction following while demonstrating attachment theory's importance for safe interactions.

\section*{Acknowledgments}
This work was supported in part by the National Natural Science Foundation of China (Grant Nos. 62276017, 62406033, U1636211, 61672081), and the State Key Laboratory of Complex \& Critical Software Environment (Grant No. SKLCCSE-2024ZX-18).

\bibliographystyle{splncs04}
\bibliography{custom}
\end{document}